\documentclass[electronics, article, submit, pdftex, moreauthors]{Definitions/mdpi} 

\preto{\abstractkeywords}{\nolinenumbers} 

\usepackage{algorithm}
\usepackage{algorithmicx}
\usepackage{algpseudocode}

\firstpage{1} 
\pubvolume{1}
\issuenum{1}
\articlenumber{0}
\pubyear{2023}
\copyrightyear{2023}
\datereceived{ } 
\daterevised{ } 
\dateaccepted{ } 
\datepublished{ } 
\hreflink{https://doi.org/} 

\Title{Exploring the Physical World Adversarial Robustness of Vehicle Detection}

\TitleCitation{Title}

\Author{Wei Jiang$^{1, \dagger}$, Tianyuan Zhang$^{1,2,*,\dagger}$\orcidA{}, Shuangcheng Liu$^{2}$, Weiyu Ji$^{1}$, Zichao Zhang$^{1}$, Gang Xiao$^{3}$}

\AuthorNames{Wei Jiang, Tianyuan Zhang, Shuangcheng Liu, Weiyu Ji, Zichao Zhang, Gang Xiao}

\AuthorCitation{Jiang, W.; Zhang, T.; Liu, S.; Ji, W.; Zhang, Z.; Xiao, G.}

\address{%
$^{1}$ \quad Information Science Academy, China Eletronics Technology Group Corporation, Beijing 100846, China\\
$^{2}$ \quad Beihang University, Beijing, China \\
$^{3}$ \quad National Key Laboratory for Complex Systems Simulation, Beijing, China\\}

\corres{Correspondence: zty929601635@gmail.com}

\firstnote{These authors contributed equally to this work.}

\abstract{Adversarial attacks can compromise the robustness of real-world detection models. However, evaluating these models under real-world conditions poses challenges due to resource-intensive experiments. Virtual simulations offer an alternative, but the absence of standardized benchmarks hampers progress. Addressing this, we propose an innovative instant-level data generation pipeline using the CARLA simulator. Through this pipeline, we establish the Discrete and Continuous Instant-level (DCI) dataset, enabling comprehensive experiments involving three detection models and three physical adversarial attacks. Our findings highlight diverse model performances under adversarial conditions. Yolo v6 demonstrates remarkable resilience, experiencing just a marginal 6.59\% average drop in average precision (AP). In contrast, the ASA attack yields a substantial 14.51\% average AP reduction, twice the effect of other algorithms. We also note that static scenes yield higher recognition AP values, and outcomes remain relatively consistent across varying weather conditions. Intriguingly, our study suggests that advancements in adversarial attack algorithms may be approaching its ``limitation''.In summary, our work underscores the significance of adversarial attacks in real-world contexts and introduces the DCI dataset as a versatile benchmark. Our findings provide valuable insights for enhancing the robustness of detection models and offer guidance for future research endeavors in the realm of adversarial attacks.}

\keyword{Adversarial Attack; Virtual Simulation; Intelligent Perception; Autonomous Driving}

\begin{document}

\section{Introduction}

In recent years, advancements in artificial intelligence (AI) technology, epitomized by deep neural networks (DNNs), have realized significant breakthroughs in areas such as computer vision \cite{krizhevsky2012imagenet,redmon2018yolov3,he2017mask}, natural language processing \cite{bahdanau2014neural}, speech recognition \cite{hinton2012deep}, and autonomous driving. These strides have ignited a transformative wave, stimulating growth in societal productivity and catalyzing progress.

Nonetheless, these deep learning methodologies encounter formidable obstacles within the complexities of real-world application scenarios. These include environmental dynamics, input uncertainties, and even potential malevolent attacks, all of which expose vulnerabilities related to security and stability. Research has indicated that deep learning can be significantly influenced by adversarial examples; through meticulous application of almost imperceptible noise, these models can be misguided into making high-confidence yet inaccurate predictions \cite{szegedy2013intriguing, goodfellow2014explaining}. This emphasizes the inherent unreliability and uncontrollability in the current generation of deep learning models. In recent years, a proliferation of adversarial attack algorithms have been introduced \cite{kurakin2018adversarial, evtimov2017robust, liu2020bias, wei2019rpc, duan2020adversarial, liu2020spatiotemporal, zhang2018camou, huang2020universal}, underscoring the threats posed by adversarial examples in the digital domain. As exploration of adversarial examples continues, it has become evident that AI systems deployed in the physical world are also susceptible to these security challenges, potentially leading to catastrophic security incidents. Therefore, research into adversarial security of deep learning in physical world applications, robustness testing of models, and ensuring the security and trustworthiness of AI systems has become an urgent imperative.

Unlike the controllable conditions in digital experiments, investigations into adversarial attacks and defenses in the physical world emphasize addressing real-world challenges due to the openness of the experimental scenarios and the variability of environmental conditions. Adversarial examples in the physical world refer to a unique type of samples created by various means such as stickers or paint that alter the features of real objects and can mislead deployed deep learning models post sampling. For instance, in autonomous driving scenarios, when executing vehicle recognition and adversarial safety testing, road conditions must first be closed off. Then, surface features are modified using coatings, stickers, and the like at specific locations on the vehicle surface. Furthermore, data captured by sensors could be error-prone due to fluctuations in weather, lighting, distance, angles, among other factors. This mode of testing not only requires significant resources but also grapples with the challenge of replicating the exact experimental environment, thereby complicating the detection of security vulnerabilities.

\begin{figure}
    \centering
    \includegraphics[width=\textwidth]{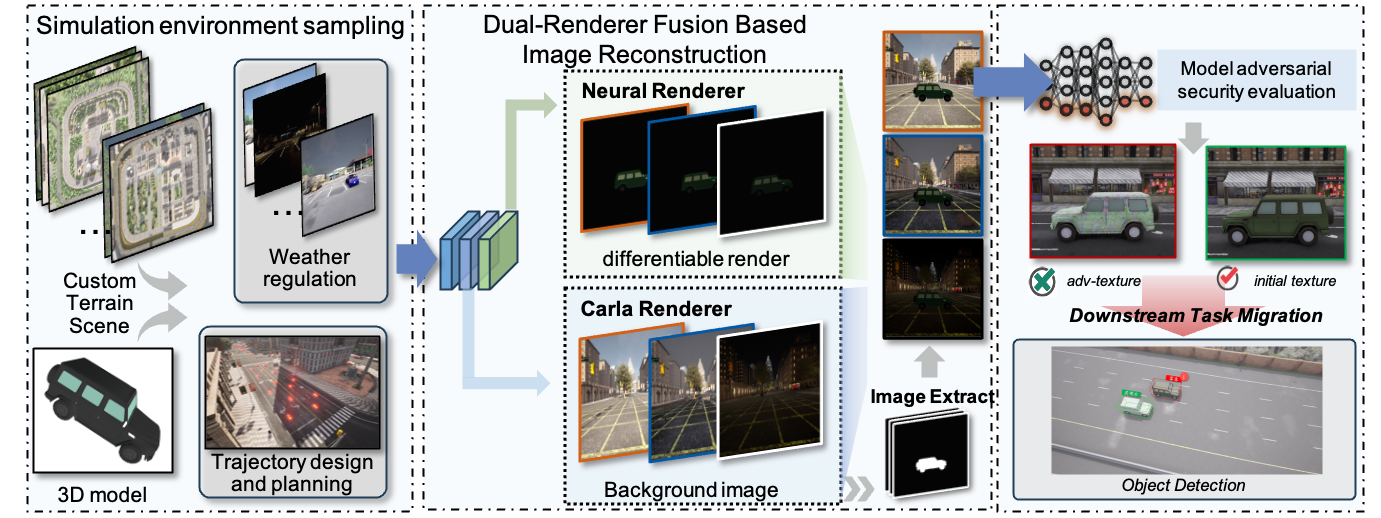}
    \caption{The framework of the entire research. This includes the process of sampling from a virtual environment, rendering with dual renderers, and model testing.}
    \label{fig:framework}
\end{figure}

To facilitate credible evaluation of deep learning models in real-world application scenarios, several adversarial attack and defense studies utilizing simulation sandboxes have been introduced \cite{zhang2018camou, wu2020physical, xiao2019meshadv}. By leveraging a physical simulation sandbox powered by a real physics engine, one can model physical scenes, construct and combine real objects, thereby enabling research into adversarial attack and defense techniques in the physical world. Research grounded in simulation sandboxes can effectively circumvent the challenges of inconvenient testing, high replication difficulty, and excessive testing costs inherent in real-world physical environments. Yet, a universally acknowledged benchmark that could guide such research is still lacking. To address this void, we introduce an instant-level scene generation pipeline based on CARLA and present the Discrete and Continuous Instant-level (DCI) dataset, which includes diverse scenarios with varying sequences, perspectives, weather conditions, textures, among others. The research framework is depicted in Figure \ref{fig:framework}. Our primary \textbf{contributions} can be distilled as follows.

\begin{itemize}
\item[$\bullet$] We present the Discrete and Continuous Instant-level (DCI) dataset, a distinct contribution that sets a benchmark for assessing the robustness of vehicle detection systems under realistic conditions. This dataset facilitates researchers in evaluating the performance of deep learning models against adversarial examples, with a specific emphasis on vehicle detection.

\item[$\bullet$] We perform a thorough evaluation of three detection models and three adversarial attack algorithms utilizing the DCI dataset. Our assessment spans various scenarios, illuminating the efficacy of these attacks under diverse conditions. This comprehensive evaluation offers insights into the performance of these models and algorithms under a range of adversarial conditions, contributing to the ongoing quest to enhance the robustness and reliability of AI systems against adversarial attacks.
\end{itemize}

\section{Related Work}
\subsection{Adversarial Attack in the Digital World}
Adversarial samples are specially designed samples that are not easily perceived by humans but can lead to erroneous judgments in deep learning models. According to the scope of the attack, adversarial attacks can be divided into two types: digital world attacks and physical world attacks.

In the digital world, adversarial attacks directly manipulate image pixels. Szegedy et al. \cite{szegedy2013intriguing} initially proposed adversarial examples, generating them through the L-BFGS method. Capitalizing on target model gradients, Goodfellow et al. \cite{goodfellow2014explaining} introduced the Fast Gradient Sign Method (FGSM) for rapid adversarial example generation. Kurakin et al.\cite{kurakin2018adversarial} enhanced FGSM, developing iterative versions, the Basic Iterative Method (BIM) and Iterative Least Likely Class Method (ILCM).  Madry et al. \cite{madry2017towards} incorporated a "Clip" function to project and added random perturbations during initialization, culminating in the widely-used Projective Gradient Descent (PGD) attack method. 

\subsection{Adversarial Attack in the Physical World}
Physical adversarial attacks often involve altering an object's visual attributes such as painting, stickers, or occlusion. They are broadly divided into two categories: (1) two-dimensional attacks (2) three-dimensional attacks.

Two-dimensional attacks are typically executed via the application of distinct patterns or stickers to the targeted objects. Sharif et al. \cite{sharif2016accessorize} deceived facial recognition systems by creating wearable eyeglass frames that mislead models in the physical world. Brown et al. \cite{brown2017adversarial} designed "adversarial patches," which are small perturbed areas that can be printed and pasted to effectively conduct an attack. Eykholt et al. \cite{eykholt2018robust} devised the Robust Physical Perturbation (RP2) method, which misguides traffic sign classifiers using occlusion textures. Thys et al. \cite{thys2019fooling} demonstrated an attack on human detection models by attaching a two-dimensional adversarial patch to the human torso. Sato et al. \cite{sato2021dirty} proposed the Dirty Road attack, misleading autonomous vehicles' perception modules by painting camouflages on lanes. Liu et al. \cite{liu2023x} proposed X-adv, implemented adversarial attacks against X-ray security inspection systems.
 
While two-dimensional physical attacks have proven effective, they are constrained by sampling angles and other conditions, thus their success in general scenarios is not guaranteed. Three-dimensional physical attacks provide a solution. Athalye et al. \cite{athalye2018synthesizing} introduced the Expectation Over Transformation (EOT) framework to create adversarial attacks on 2D images and 3D objects. In contrast, Maesumi et al. \cite{maesumi2021learning} proposed a 3D-to-2D adversarial attack method, using structured patches from a reference mannequin, with adaptable human postures during training.

Moreover, three-dimensional attacks are also executed using simulation environments. Zhang et al. \cite{zhang2018camou} and Wu et al. \cite{wu2020physical} utilized open-source virtual simulation environments for their optimized adversarial attacks. Wang et al. \cite{wang2021dual} introduced Dual Attention Suppression (DAS) to manipulate attention patterns in models. Zhang et al. \cite{kato2018neural} developed the Attention on Separable Attention (ASA) attack, enhancing the effectiveness of adversarial attacks.

Given the emergence of numerous adversarial attack and defense studies , the establishment of a benchmark for a comprehensive security analysis of these algorithms becomes imperative.

\subsection{Adversarial Robustness Benchmark}
Several physical-world adversarial example generation methods have been proposed and demonstrated to be effective \cite{liu2019perceptual, liu2020spatiotemporal, liu2020bias, wang2021universal, wang2022defensive, liu2022harnessing, liu2023exploring, guo2023towards}. However, they use different dataset for evaluation, which makes it difficult to conduct a comprehensive evaluation. To address this issue, several benchmarks have been proposed, including those by Dong  \cite{dong2020benchmarking} and Liu  \cite{liu2021training}. Tang \cite{tang2021robustart} proposed the first unified Robustness Assessment Benchmark, RobustART, which provides a standardized evaluation framework for adversarial examples.

In the virtual simulation environment, several adversarial attack algorithms for vehicle recognition scenarios have been proposed\cite{zhang2018camou, wang2021dual, zhang2022transferable, wang2022fca} and shown to be effective. The CARLA simulator\cite{dosovitskiy2017carla} has been widely used in these studies due to its versatility and availability. However, the lack of a unified evaluation benchmark makes it difficult to compare and analyze the results. Establishing a benchmark is essential to promote the development of robust vehicle detection models.

\subsection{Virtual Environment of Vehicle Detection}
A series of vehicle detection-related simulators have been proposed. Simulators developed based on the Unity engine, such as LGSVL\cite{rong2020lgsvl}, and those developed based on the Unreal engine, such as Airsim\cite{shah2018airsim} and CARLA\cite{dosovitskiy2017carla}, all support camera simulation. Among them, the Airsim simulator focuses more on drone-related research, while compared with LGSVL, current research on adversarial security is more focused on the CARLA simulator\cite{wang2021dual,zhang2022transferable,wang2022fca}. CARLA is equipped with scenes and high-precision maps made by RoadRunner, and provides options for map editing. It also supports environment lighting and weather adjustments, as well as the simulation of pedestrian and vehicle behaviors.

Based on the above exploration, this study intends to use the CARLA autonomous driving simulator as the basic simulation environment to carry out research on the security analysis of autonomous driving intelligent perception algorithms.
\section{DCI Dataset: Instant-level Scene Generation and Design}
\subsection{Neural 3D Mesh Render Technology}
Neural 3D Mesh Renderer \cite{kato2018neural} is an image rendering technique based on deep learning that leverages trained neural networks to produce high-quality images. Traditional image rendering approaches typically require the manual definition of intricate rendering rules and optical models, using rasterization and shading techniques to generate realistic images. In contrast, neural rendering methods simplify this process by employing deep neural networks to automatically learn these rules and models. Additionally, these techniques ensure the traceability of the gradient of the training textures during the rendering process, facilitating the training and evaluation of adversarial attack and defense samples.

\subsection{Dual-Renderer Fusion Based Image Reconstruction}

\begin{table}[] 
\caption{Illustration of parameter transfer between renderers. The parameters fall into two categories: 1) Position coordinates and 2) Environmental parameters. Each category plays a vital role in enhancing the realism and fidelity of the rendered images.}
\newcolumntype{C}{>{\centering\arraybackslash}X}

\renewcommand{\arraystretch}{1.2} 
\begin{tabularx}{\textwidth}{CC}
\toprule
\textbf{Categories}	& \textbf{Parameter Name}	\\ \hline
\multirow{3}{*}{\textbf{Positional coordinate $P_{co}$}} & $Model_{angle}$  \\ \cline{2-2} 
 & $Camera_{up}$ \\ \cline{2-2} 
 & $Camera_{direction}$ \\ \hline
\multirow{5}{*}{\textbf{Environmental parameters $P_{en}$}} & $Intensity_{direction}$ \\ \cline{2-2} 
 & $Intensity_{directional}$ \\ \cline{2-2} 
 & $Color_{ambient}$ \\ \cline{2-2} 
 & $Color_{directional}$ \\ \cline{2-2} 
 & $Direction$ \\
\bottomrule
\end{tabularx}
\renewcommand{\arraystretch}{1.0} 
\label{tab:parameters}
\end{table}

In this study, we have developed an instance-level scene generation approach that effectively combines the CARLA simulator and the Neural renderer. The CARLA simulator, an image-based rendering tool, provides high fidelity and precision in detail, but is hindered by its inability to differentiate textures generated during rendering, which hampers adversarial sample generation and gradient-based optimization. In contrast, neural renderers bypass this limitation, preserving gradient traceability during the rendering process which is essential for creating adversarial samples and enhancing adversarial attacks. This fusion of rendering tools not only facilitates the production of highly realistic scene imagery but also supports the generation and optimization of adversarial attack methods by utilizing gradient information from the rendering process.


In previous studies, the only parameter passed between the two renderers was the positional coordinate $P_{co}$. While this method ensures consistency in the model's appearance pre and post-rendering, it disregards the influence of environmental factors such as lighting changes on the final render quality. Consequently, in the synthesized image, the rendered 3D model appears with the same lighting effect under different environmental conditions (like sunny, rainy, night, etc.), which significantly impairs the image's realism. To address this issue, we introduced environmental parameters $P_{en}$ as additional transfer parameters to minimize the difference in lighting between the two renderers, thus enhancing the realism of the render. By integrating the CARLA simulator and the neural renderer in this manner, we have successfully built an instance-level scene generation method that guarantees scene realism while also supporting the training of adversarial textures. The parameters passed are shown in 
Table \ref{tab:parameters}.

\begin{figure}
    \centering
    \includegraphics[width=\textwidth]{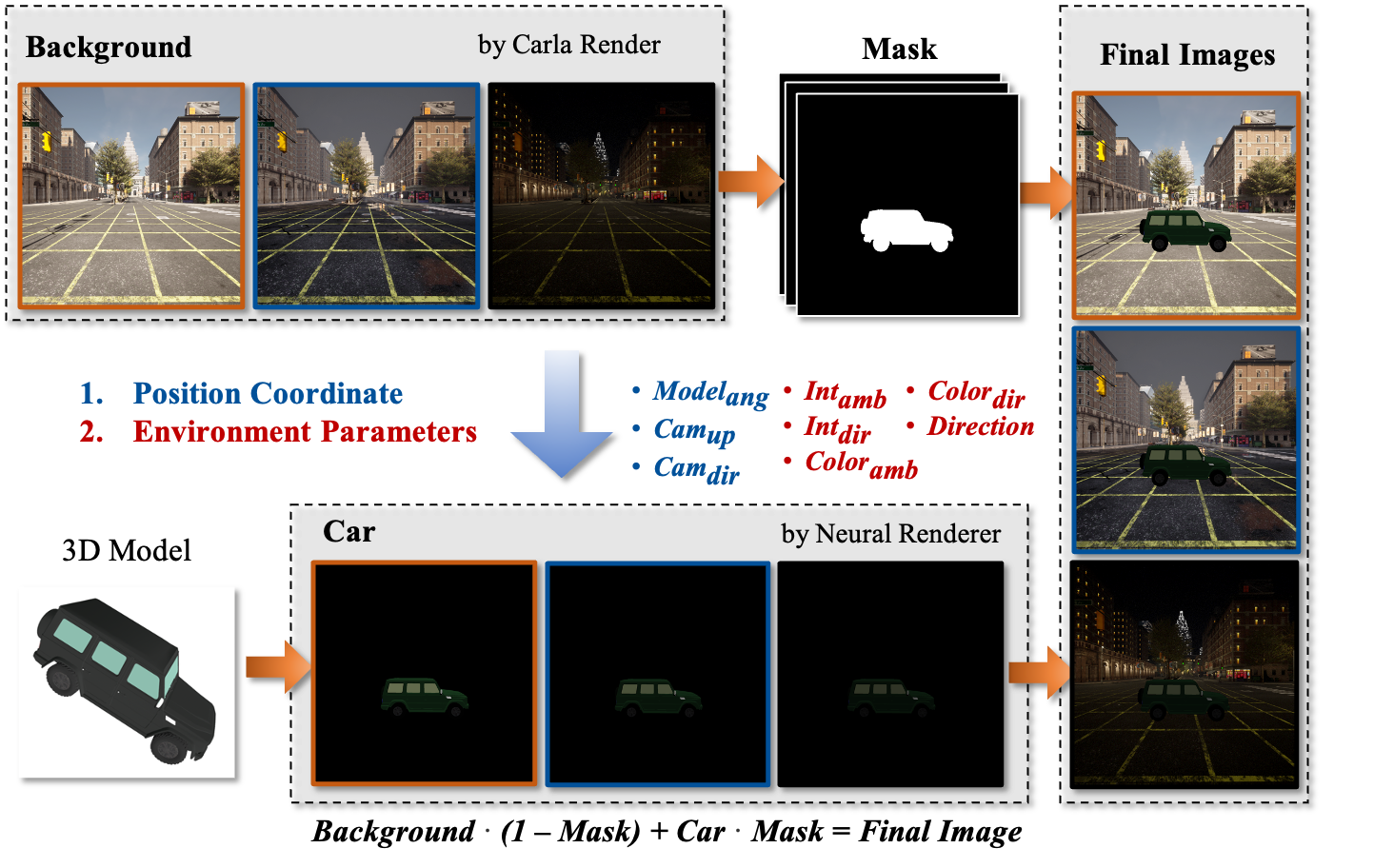}
    \caption{The pipeline of dual-renderer fusion based image reconstruction}
    \label{fig:SceneGeneration}
\end{figure}

Specifically, we use the CARLA simulator to first generate the \textbf{\textit{Background}} image and obtain the position coordinates $P_{co}$ and environment parameters $P_{en}$ using the simulator's built-in sensor. Next, we transfer $P_{co}$ and $P_{en}$ to the neural renderer. The Neural renderer then loads the 3D model and uses the received parameters to generate the \textbf{\textit{Car}} image. During the rendering process, we adjust the relevant settings of the neural renderer according to the sampling environment in CARLA to narrow the gap between the two renderers. We then use a \textbf{\textit{Mask}} to extract the background image and vehicle image respectively. After completing the pipeline, we obtain an instant-level scene. The framework of scene generation is shown in Figure \ref{fig:SceneGeneration}.

By introducing environmental parameters $P_{en}$, we have successfully enhanced the overall quality of the scene generation process and maintained consistency between the renderers under varying lighting conditions. This approach lays a significant foundation for analyzing the robustness and security of deep learning models under various environmental conditions. We believe that this instance-level scene generation method can provide more comprehensive and reliable support for the adversarial safety testing and evaluation of autonomous driving systems.

\subsection{Connected Graphs based Case Construction}
When constructing scenario execution cases, understanding certain fundamental concepts is crucial. In the CARLA simulator, Actors refer to objects that can be arbitrarily positioned, set to follow motion trajectories, and perform actions. These include vehicles, pedestrians, traffic signs, traffic lights, sensors, and more. These actors play various roles in the simulation scenario, and their interactions significantly influence the overall simulation process. Notably, CARLA's sensors, such as RGB cameras and instance segmentation cameras, can be attached to other Actors for data collection. By strategically positioning the Actor and determining its action trajectory, a broad range of scenarios and execution instances can be generated for testing various algorithms and models.

The original approach to generating and setting Actors in the CARLA simulator involves manually determining each Actor's position, speed, displacement distance, and steering angle. However, this method has issues such as slow generation speed and lack of realistic simulation effects, necessitating improvements. As a solution, we utilized an optimization method based on Connected Graph generation. This method automatically generates information such as the location, number, and action track of Actors via a program, enabling quick construction of numerous execution instances. Specifically, the CARLA built-in map contains several "spawn points." By setting these spawn points on the running path and incorporating the A* shortest path generation algorithm, we can quickly generate and realistically simulate Actor trajectories. The algorithm's pseudocode is presented in Algorithm \ref{alg:shortestpath}.

\begin{algorithm}[H]
\caption{Connected Graphs based Case Construction}
\label{alg:shortestpath}
\begin{algorithmic}[1]
\State \textbf{/* Initialization */}
\State Initialize empty set $OpenList$ and add $startNode$ to it
\While {$OpenList \neq \varnothing$}
    \State $currentNode \leftarrow \arg\min_{node \in OpenList} f(node)$
    \State $childSet \leftarrow$ \{Children of $currentNode$ that are valid and not visited\}
    \For {each $childNode$ in $childSet$}
        \State Calculate $f(childNode)$ considering $currentNode$ as parent
        \If {$f(childNode)$ can be improved}
            \State Update parent of $childNode$ to $currentNode$
        \EndIf
    \EndFor
    \State Remove $currentNode$ from $OpenList$
    \If {$endNode \in childSet$}
        \State break
    \EndIf
\EndWhile
\State $shortestPath \leftarrow$ Trace back from $endNode$ to $startNode$
\State \Return $shortestPath$
\end{algorithmic}
\end{algorithm}

\subsection{Composition of the DCI Dataset}
\begin{figure}
    \centering
    \includegraphics[width=0.95\textwidth]{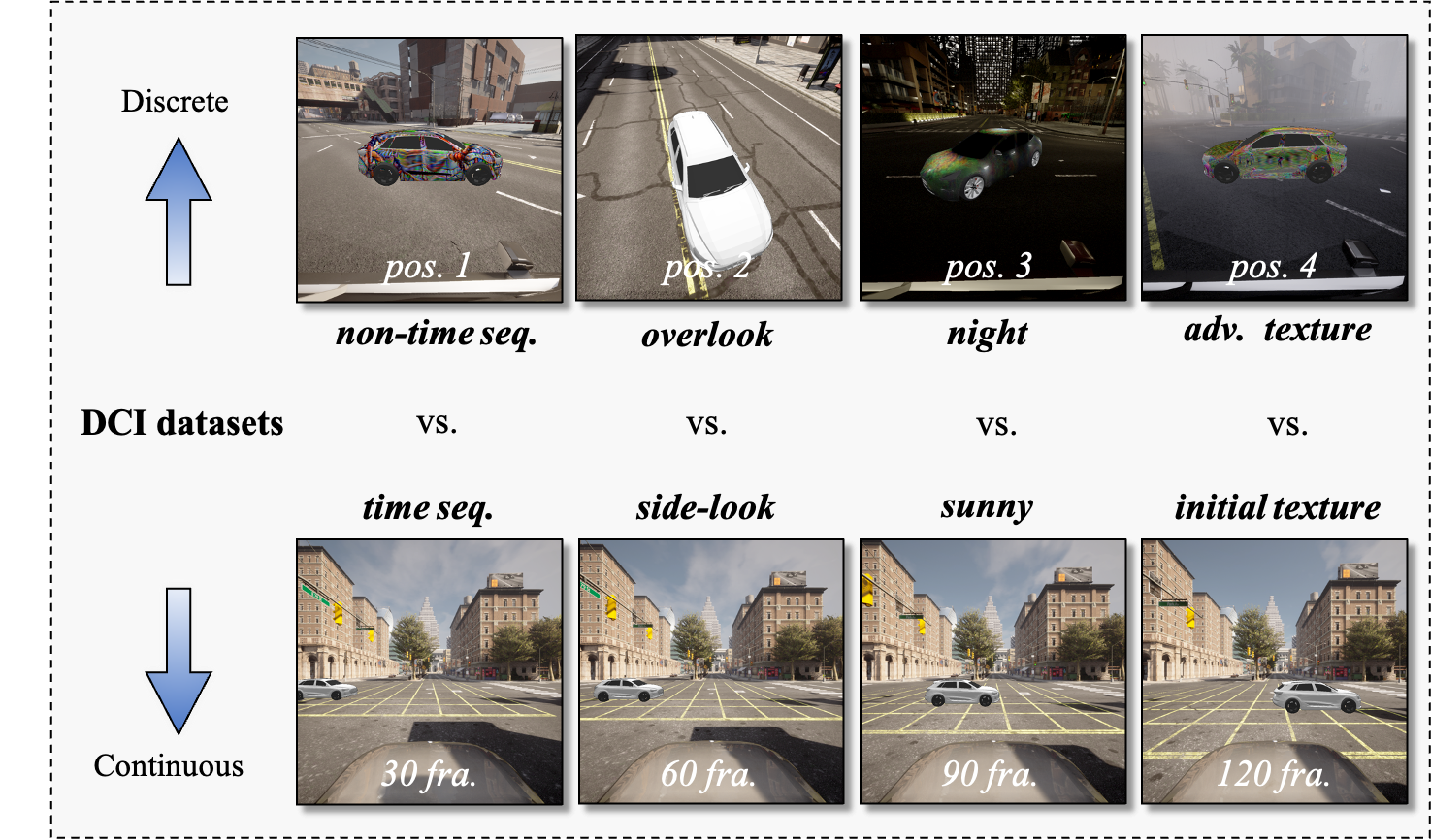}
    \caption{The Discrete and Continuous Instant-level Dataset (DCI): the discrete part aims to provide all-round coverage, while the continuous part is designed to test specific scenarios in greater depth.}
    \label{fig:DCIdataset}
\end{figure}

The Discrete and Continuous Instant-level (DCI) dataset is designed to evaluate the performance of vehicle detection models in diverse scenarios. It can be divided into two parts that focus on different aspects. Figure \ref{fig:DCIdataset} illustrates various components of the DCI dataset.

The \textbf{Continuous} part of the DCI dataset comprises 7 typical scenes, each describing a real-life scenario that is widely used. To address the issue of irregular data distribution, we employed a fixed viewpoint approach to address issues of uneven data distribution and insufficient scene representation. This approach includes the driver's perspective and monitor view. The driver's viewpoint simulates the field of view of an on-road driver, while the drone viewpoint offers a comprehensive bird's-eye view of the scene. Lastly, the surveillance viewpoint resembles that of a fixed surveillance camera. This multi-viewpoint strategy broadens our data collection scope, significantly enhancing the dataset's quality and diversity. To expand the coverage, we chose three different weather conditions to generate the dataset: \textit{ClearNoon}, \textit{ClearNight}, and \textit{WetCloudySunset}. This part of data set involves seven angles, distances and more than 2000 different positions.

The \textbf{Discrete} part of the DCI dataset aims to extend coverage by widely selecting parameters such as map locations, sampling distances, pitch angles, and azimuth angles, encompassing various road types and topological structures. We traverse road locations on the map while fine-tuning lighting angles and intensities to simulate variations in illumination under different times and weather conditions. Moreover, we adjust environmental conditions like haze and particle density, thereby enhancing the dataset's authenticity and diversity. This segment includes 40 angles, 15 distances, and over 20,000 distinct locations. The composition of the DCI dataset is shown in Table \ref{tab:DCIdataset}.

\begin{table}[] 
\caption{Overview of the DCI dataset.\label{tab:DCIdataset}}
\newcolumntype{C}{>{\centering\arraybackslash}X}
\begin{tabularx}{\textwidth}{CC}
\toprule
\textbf{Scene Name}	& \textbf{Scene Description (Perspective)}	\\
\midrule
Overall\textsuperscript{1}		& Random (Random)		\\
Traffic Circle		& Driving in the center of the road (Monitor)\\
Parking Lot		& Exiting from a parking lot (Driver)\\
Stationary A		& Stationary observation (Driver)\\
Straight A		& Driving straight on a road (Driver)\\
Turning A		& Turning at an intersection (Driver)\\
Stationary B		& Stationary observation (Driver)\\
Straight B		& Driving straight on a road (Driver)\\
\bottomrule
\end{tabularx}
\noindent{\footnotesize{\textsuperscript{1} This is the discrete part, while the others are all continuous parts.}}
\end{table}
\section{Experiments and Evaluations}

\subsection{Experiment Settings}  
\textbf{Adversarial Attack Algorithm.} We employed three algorithms to generate adversarial examples: DAS algorithm \cite{wang2021dual}, FCA algorithm \cite{wang2022fca}, and ASA algorithm \cite{zhang2022transferable}. These algorithms were carefully chosen based on their proven effectiveness in generating adversarial examples and their compatibility with our proposed method. The adversarial texture was trained on the discrete dataset mentioned earlier, utilizing 1 epoch, a batch size of 1, and an iteration step size of 1e-5. 

\textbf{Vehicle 3D Model.} We used the Audi E-Tron, a commonly used 3D model in previous studies, for our experiments. The model comprises 13,449 vertices, 10,283 vertex normals, 14,039 texture coordinates, and 23,145 triangles.

\textbf{Vehicle Detection Algorithm.} We evaluated the proposed method on three popular object detection algorithms: YOLO v3 \cite{redmon2018yolov3} YOLO v6\cite{li2022yolov6} and Faster R-CNN \cite{ren2015faster}. By selecting both single-stage and two-stage typical algorithms, we investigated the capability of the attack algorithm in the real world. The target class we chose is the car. We used the Average Precision (AP) as the evaluation metric to measure the performance of the detection algorithm on the test dataset.

\subsection{Analysis of Experimental Results in Discrete Part}

\begin{table}[H] 
\caption{Accuracy of Vehicle Detection (AP) in Continuous Scenarios.\label{tab:overall}}
\newcolumntype{C}{>{\centering\arraybackslash}X}
\begin{tabularx}{\textwidth}{CCCC}
\toprule
\textbf{Texture Type} & \textbf{AP@yolov3(\%)} & \textbf{AP@yolov6(\%)} & \textbf{AP@FRCNN(\%)}\\
\midrule
initial texture		& 65.37 & \textbf{73.39} & 56.81\\
ASA adv-texture		& 41.59 & \textbf{65.68} & 44.76\\
DAS adv-texture		& 57.39 & \textbf{65.66} & 50.49\\
FCA adv-texture		& 456.8 & \textbf{68.44} & 47.21\\
\bottomrule
\end{tabularx}
\end{table}

We initially selected the Overall Coverage scenario for analysis, which allows for a comprehensive performance assessment under various conditions. We use the "@" symbol to represent the corresponding model, as shown in Table \ref{tab:overall}. Under original texture conditions, the yolov6 model exhibited the highest AP value, reaching 73.39\%. The yolov6 model was closely followed by the yolov3 model, with an AP of 65.37\%. Meanwhile, the Faster RCNN model had the lowest AP value, at just 56.81\%.

Thus, under conditions free from adversarial attacks, the order of \textbf{detection accuracy rates} is as follows: \textbf{yolov6 > yolov3 > Faster RCNN}.

Switching vehicle textures to adversarial forms resulted in notable shifts in model detection accuracy. Under ASA adversarial texture, the performance of the yolov3 model was notably diminished, registering an AP of 41.59\%, less than the Faster RCNN model's 44.76\% AP. This anomaly may stem from the significant impact of the ASA adversarial texture on the yolov3 model.

Contrastingly, in DAS and FCA adversarial scenarios, the yolov3 model outperformed Faster RCNN, recording APs of 57.39\% and 56.8\%, compared to Faster RCNN's 50.76\% and 47.21\%. This highlights yolov3's relative resilience and stability under these adversarial conditions.

To assess the effectiveness of adversarial texture attacks, merely observing the average precision (AP) values can be insufficient as these can be influenced by a myriad of factors. To more precisely evaluate the attack effects, we considered the decline in AP. Thus, we calculated the average AP drop rates under adversarial texture conditions for various object detection models, as presented in Table \ref{tab:overallDecline}.

\begin{figure}
    \centering
    \includegraphics[width=0.95\textwidth]{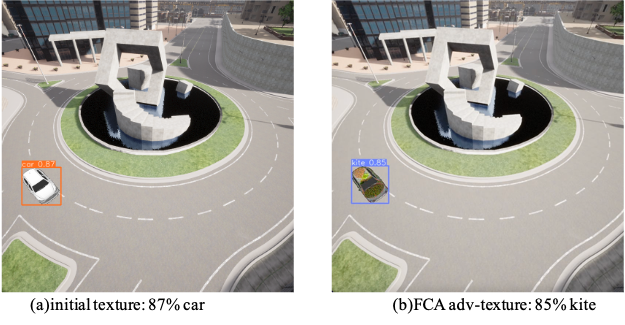}
    \caption{Illustration of yolov3 model's performance under original and FCA adversarial textures. The model correctly identifies a car with 87\% confidence under original texture but misclassifies it as a kite with 85\% confidence under the FCA adversarial texture.}
    \label{fig:compare}
\end{figure}

Following the implementation of adversarial perturbations, the mean decrease in AP was 13.44\%, 6.79\%, and 9.32\% for yolov3, yolov6, and Faster RCNN models respectively. Notably, the yolov6 model demonstrates the highest resilience with the least AP decrease, while the yolov6 model has the weakest with the most significant drop. The Faster RCNN model presents good robustness but is slightly behind the yolov6 model. To visually demonstrate attack effects, we utilized the yolov3 model on a frame rendered with original and FCA adversarial textures (Figure \ref{fig:compare}). 

In summary, when subjected to adversarial texture attacks, the yolov6 model exhibits superior robustness, while yolov3 presents less resilience. The \textbf{robustness ranking} is as follows: \textbf{yolov6 > Faster RCNN > yolov3}.

\begin{table}[H] 
\caption{Accuracy Decline of Vehicle Detection (AP) in Continuous Scenarios.\label{tab:overallDecline}}
\newcolumntype{C}{>{\centering\arraybackslash}X}
\begin{tabularx}{\textwidth}{CCCC}
\toprule
\textbf{Texture Type} & \textbf{AP@yolov3(\%)} & \textbf{AP@yolov6(\%)} & \textbf{AP@FRCNN(\%)}\\
\midrule
ASA adv-texture		& \textbf{23.78} & 7.71 & 12.05\\
DAS adv-texture		& \textbf{7.98} & 7.74 & 6.32\\
FCA adv-texture		& 8.57 & 4.95 & \textbf{9.6}\\
\bottomrule
\end{tabularx}
\end{table}

\subsection{Analysis of Experimental Results in Continuous Part}

Upon analysis of the Overall coverage scenario, we further delve into various subdivided scenarios to examine the models' recognition performance in diverse environments, with specific results presented in Table \ref{tab:continuous}. Through these more granulated scenario experiments, we have observed some differing results.

Primarily, the yolov6 model still exhibits the highest recognition accuracy across most scenarios. This indicates that the yolov6 model has superior performance and can maintain high accuracy across a multitude of subdivided scenarios.

\begin{table}[] 
\caption{Accuracy  of Vehicle Detection (AP) in Continuous Scenarios.\label{tab:continuous}}
\newcolumntype{C}{>{\centering\arraybackslash}X}
\begin{tabularx}{\textwidth}{CCCC}
\toprule
\textbf{Scene}  & \textbf{AP@yolov3(\%)} & \textbf{AP@yolov6(\%)} & \textbf{AP@FRCNN(\%)} \\
\midrule
Traffic Circle      & 63.36     & \textbf{86.95}     & 19.43    \\
Parking Lot         & 23.03     & 32.33     & \textbf{33.64}    \\
Stationary A        & 66.27     & \textbf{68.99}     & 67.89    \\
Straight Through A  & 80.93     & \textbf{82.49}     & 81.86    \\
Turning Left A      & 48.02     & \textbf{64.34}     & 18.26    \\
Stationary B        & 98.81     & \textbf{100}       & 100      \\
Straight Through B  & 76.72     & \textbf{78.68}     & 76.57    \\
\bottomrule
\end{tabularx}
\end{table}

Interestingly, within the overall poorer-performing Faster RCNN model, we found that it achieved the highest accuracy among the three models in the ``Parking Lot'' and ``Stationary B'' scenarios. This implies that while the Faster RCNN model can perform well in specific scenarios, it tends to be unstable in others.

In our scene-specific tests, we evaluated the models' AP decline under adversarial attacks, revealing the variance in algorithm performance across diverse scenes (Table \ref{tab:continuous_decline}). The yolov3 model demonstrated the most significant AP decline, often exceeding 20\%. Conversely, yolov6 and Faster RCNN showed a more stable AP decline, consistently under 20\%. This implies that model robustness varies across scenes, with yolov3 particularly needing additional optimization for complex environments, while yolov6 and Faster RCNN display superior robustness.

\begin{table}[] 
\caption{Accuracy Decline of Vehicle Detection (AP) in Continuous Scenarios.\label{tab:continuous_decline}}
\newcolumntype{C}{>{\centering\arraybackslash}X}
\begin{tabularx}{\textwidth}{CCCC}
\toprule
\textbf{Scene}  & \textbf{AP@yolov3(\%)} & \textbf{AP@yolov6(\%)} & \textbf{AP@FRCNN(\%)} \\
\midrule
Traffic Circle      & \textbf{13.44}     & 6.80     & 9.32    \\
Parking Lot         & \textbf{20.66}     & 17.89    & 7.95    \\
Stationary A        & 10.02     & 14.70    & \textbf{19.23}   \\
Straight Through A  & \textbf{25.98}     & 12.82    & 10.19   \\
Turning Left A      & \textbf{7.15}      & 0.46     & 6.36    \\
Stationary B        & -7.41     & -7.01    & \textbf{2.86}   \\
Straight Through B  & \textbf{25.72}     & 6.21     & 7.12    \\
\bottomrule
\end{tabularx}
\end{table}

\begin{figure}
    \centering
    \includegraphics[width=0.95\textwidth]{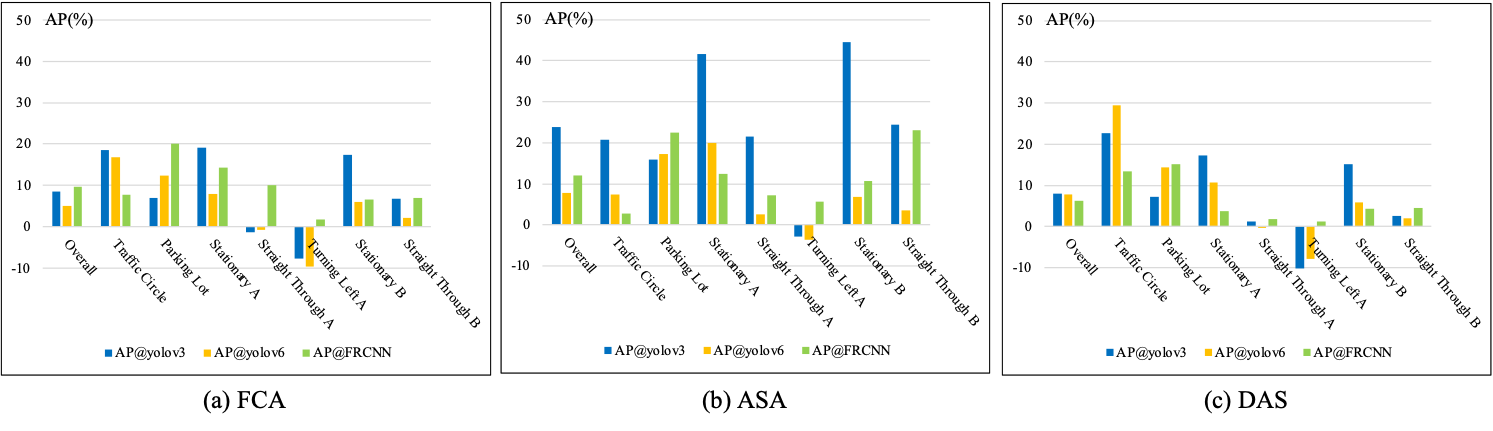}
    \caption{Average Decrease in AP for Detection Models Under Attack in Various Scenarios}
    \label{fig:accDecline_three}
\end{figure}

Upon detailed examination of various adversarial attack algorithms, as illustrated in Figure \ref{fig:accDecline_three}, we observe a pronounced drop in the AP for the yolov3 model under the ASA attack, underlining its weakest robustness against this specific adversarial scenario. Interestingly, the AP decrease across other adversarial attack methods does not show significant discrepancies for the remaining models. This observation suggests that the impact of different adversarial attacks on target detection models varies significantly. In particular, the yolov3 model exhibits a substantial decrease in performance under ASA, resulting in a considerable drop in AP. However, in other adversarial scenarios, all three models showcase comparable levels of robustness, indicating a relatively strong resistance to adversarial texture attacks.

Thus, under adversarial attacks in specific scenarios, the robustness of the three object detection algorithms is ranked as follows: \textbf{yolov6 > Faster RCNN > yolov3}, which aligns with the results observed in overall scenarios.

\subsection{Analysis of Specific Scenarios}
\begin{figure}
    \centering
    \includegraphics[width=0.95\textwidth]{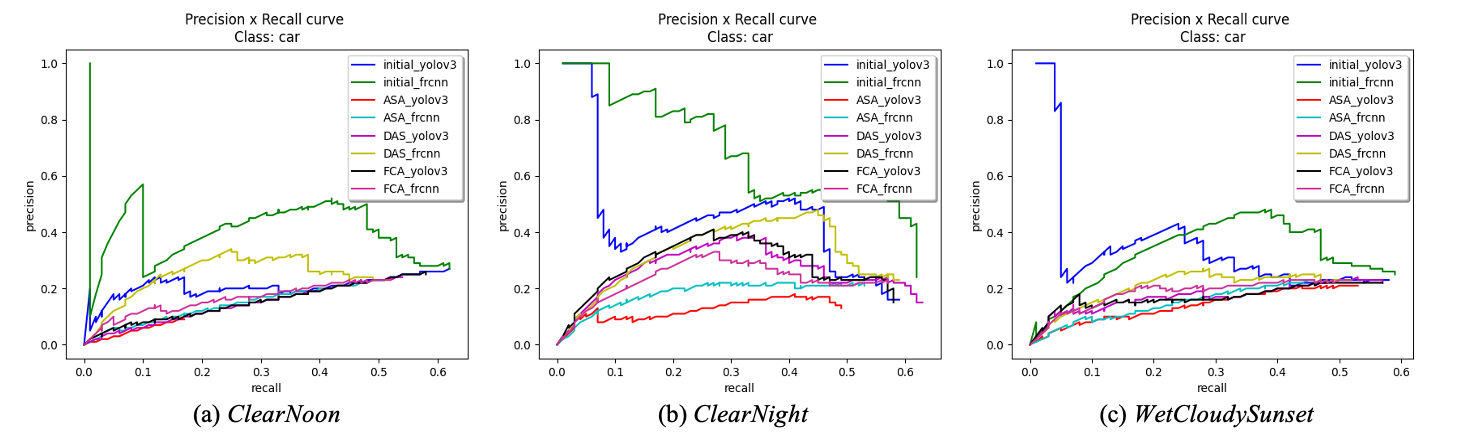}
    \caption{The Precision-Recall chart illustrates the Park Lot scenario in three different weather conditions, demonstrating a similar distribution of values.}
    \label{fig:three_graphs}
\end{figure}

We elected to further analyze the Parking Lot scenario. Figure \ref{fig:three_graphs} presents the Precision-Recall curves corresponding to different adversarial textures under the yolov3 and Faster RCNN models. As expected, the two lines with the highest values correspond to the initial textures. Interestingly, under other adversarial texture conditions, while there are numerical differences in the data distributions, they exhibit similar trends and patterns. This is also observed in the analysis of the Straight Through A and Straight Through B scenarios.

Considering that the attack magnitude was unrestricted, this implies that there may be a common ``limiting factor'' among different attacks, rendering their effects similarly to a certain extent. The development of attack algorithms might progressively converge to this ``limit'' to implement more effective attacks under different scenarios and conditions. This finding has significant implications for understanding the nature of adversarial attacks and their impact in practical applications, and could guide future research direction in the realm of object detection and adversarial attacks.

\section{Conclusion}
In the experiment, Yolo v6 showed the strongest resistance to attacks with an average AP drop of only 6.59\%. ASA was the most effective attack algorithm, reducing AP by an average of 14.51\%, twice that of other algorithms. Static scenes had higher recognition AP, and results in the same scene under different weather conditions were similar. Further improvement of adversarial attack algorithms may be reaching the ``limitation''.

\vspace{6pt} 







\authorcontributions{Conceptualization, W.Jiang and T.Z.; methodology, W.Jiang and T.Z.; validation, W.Ji and Z.Z.; data curation, W.Ji; writing---original draft preparation, T.Z.; writing---review and editing, W.Jiang. and G.X.; visualization, Z.Z. and G.X. All authors have read and agreed to the published version of the manuscript.}

\funding{This research received no external funding.}

\dataavailability{The data presented in this study are available upon request from the corresponding author. The data are not publicly available due to further research plans.} 

\conflictsofinterest{The authors declare no conflict of interest.} 

\begin{adjustwidth}{-\extralength}{0cm}

\reftitle{References}


\bibliography{egbib}

\PublishersNote{}
\end{adjustwidth}
\end{document}